\documentclass[letterpaper, 10 pt, conference]{ieeeconf}  

\IEEEoverridecommandlockouts                              

\usepackage[table]{xcolor}
\usepackage{graphicx}
\usepackage{booktabs}
\usepackage{multirow}
\usepackage{array}
\usepackage{siunitx}
\usepackage{adjustbox}
\usepackage{subcaption}
\usepackage{amssymb}
\usepackage{amsmath}
\usepackage{diagbox}
\usepackage{makecell}
\usepackage[flushleft]{threeparttable}
\usepackage[compact]{titlesec}
\titlespacing{\section}{0pt}{*0.8}{*0.8}
\titlespacing{\subsection}{0pt}{*0.6}{*0.6}
\titlespacing{\subsubsection}{0pt}{*0.6}{*0.6}

\title{\LARGE \bf
BEV-DWPVO: BEV-based Differentiable Weighted Procrustes for Low Scale-drift Monocular Visual Odometry on Ground
}

\author{Yufei Wei$^{1}$, Sha Lu$^{1}$, Wangtao Lu$^{1}$, Rong Xiong$^{1}$ and Yue Wang$^{1\dagger}$%
\thanks{*This work was not supported by any organization}
\thanks{$^{1}$Laboratory of Industrial Control and Technology, and the Institute of Cyber-Systems and Control, Zhejiang University, Hangzhou, 310058, China.}
\thanks{$^{\dagger}$Corresponding author wangyue@iipc.zju.edu.cn.}
}

\begin{document}

\maketitle
\thispagestyle{empty}
\pagestyle{empty}

\begin{abstract}

Monocular Visual Odometry (MVO) provides a cost-effective, real-time positioning solution for autonomous vehicles. However, MVO systems face the common issue of lacking inherent scale information from monocular cameras. Traditional methods have good interpretability but can only obtain relative scale and suffer from severe scale drift in long-distance tasks. Learning-based methods under perspective view leverage large amounts of training data to acquire prior knowledge and estimate absolute scale by predicting depth values. However, their generalization ability is limited due to the need to accurately estimate the depth of each point.
In contrast, we propose a novel MVO system called BEV-DWPVO. \textcolor{black}{Our approach leverages the common assumption of a ground plane, using Bird's-Eye View (BEV) feature maps to represent the environment in a grid-based structure with a unified scale. This enables us to reduce the complexity of pose estimation from 6 Degrees of Freedom (DoF) to 3-DoF. Keypoints are extracted and matched within the BEV space, followed by pose estimation through a differentiable weighted Procrustes solver. The entire system is fully differentiable, supporting end-to-end training with only pose supervision and no auxiliary tasks. We validate BEV-DWPVO on the challenging long-sequence datasets NCLT, Oxford, and KITTI, achieving superior results over existing MVO methods on most evaluation metrics.}

\end{abstract}

\section{Introduction}

Monocular Visual Odometry (MVO) is widely used in ground vehicles, such as logistics robots and autonomous driving cars, due to its ease of deployment and low hardware cost. However, MVO faces the critical challenge of scale drift in long-distance scenarios due to the lack of depth information, limiting its applications.

Traditional methods, such as direct \cite{engel2014lsd, engel2017direct}, feature-based \cite{campos2021orb}, and semi-direct methods \cite{forster2014svo}, address depth uncertainties during initialization by establishing a relative scale from the first few frames as a global reference. However, binding the global scale to initial motions causes scale drift over time due to calibration errors, feature mismatches, and motion blur.

\begin{figure}[t]  
\centering
\includegraphics[width=0.48\textwidth]{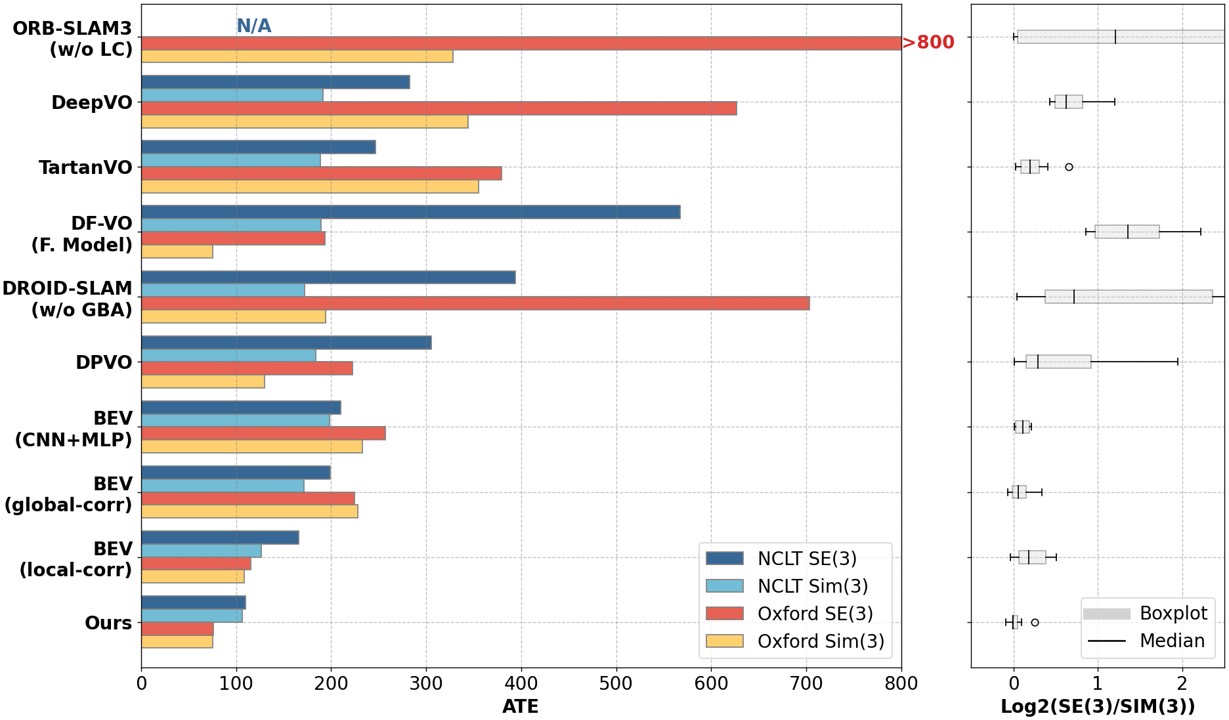}
\caption{ATE metrics for different methods on NCLT and Oxford datasets. We compare the ratio of ATE metrics for different methods under SE(3) alignment and Sim(3) alignment using the composite metric \( \log_2(\text{SE(3)}/\text{Sim(3)}) \). This comparison demonstrates the scale drift of different methods. Our method uses BEV representation and differentiable weighted Procrustes, achieving the best scale consistency and the lowest average ATE metrics across both datasets.}
\label{fig:ATE}
\vspace{-0.5cm}
\end{figure}

Learning-based methods leverage deep networks to reduce scale drift. Early works \cite{konda2015learning, wang2017deepvo} extract features from perspective images to regress poses but show limited performance. Later approaches combine deep learning’s modeling strength with traditional methods’ interpretability. For instance, \cite{li2018undeepvo, wang2021tartanvo, zhan2021df, teed2021droid, teed2024deep} employ supervised learning for optical flow and depth estimation, integrating multi-view geometry and backend optimization for improved results.
Self-supervised methods \cite{bian2019unsupervised, li2020self} train monocular depth prediction without additional supervision. However, as \cite{zhan2021df} shows, self-supervised methods still face scale ambiguity and tend to underperform when compared to supervised methods.

Recently, based on the common ground-plane assumption in autonomous driving scenarios \cite{unger2024multi}, Bird's-Eye View (BEV) perception has gained attention for tasks like 3D detection and segmentation, showing advantages over perspective view methods. Some methods \cite{ross2022bev, zhang2022bev, li2023occ} have used BEV segmentation results to achieve odometry tasks. \textcolor{black}{However, existing methods rely on segmentation supervision, leaving it unclear whether the reduced scale drift in odometry tasks is primarily due to the grid-based structure with a unified scale in the BEV representation or the segmentation applied within the BEV framework.} To create a BEV-based MVO without relying on side-tasks like segmentation, an intuitive idea is to use Convolutional Neural Networks (CNNs) to extract features or compute local cross-correlation between two BEVs for motion information, followed by Multi-Layer Perceptrons (MLPs) regression for pose estimation \textcolor{black}{\cite{wei2024bev}}. \textcolor{black}{However, recent visual odometry works have shown that incorporating geometric constraints into pose estimation yields better generalization compared to direct regression approaches \cite{zhan2021df, teed2021droid}.}

Based on these considerations, we propose BEV-DWPVO, an MVO method that leverages BEV representation and a carefully crafted pipeline for pose estimation based on two-frame BEV feature maps, trained exclusively through pose supervision. Our framework includes a vision PV-BEV encoder similar to the Lift, Splat, Shoot (LSS) \cite{philion2020lift} architecture, a UNet-like \cite{ronneberger2015u} decoder, a differentiable weighted keypoint extraction and matching module, and a differentiable weighted Procrustes pose solver \cite{gower1975generalized}.

\textcolor{black}{We evaluate our method on three datasets, comparing it against traditional approaches, learning-based perspective view methods, and three customized BEV-based methods. This comparison isolates the impact of "BEV representation" as well as "BEV-based feature extraction, matching, and weighted Procrustes pose solver." On the more challenging NCLT and Oxford datasets, our method achieves the best average RTEs of 8.67\% and 6.92\%, and RREs of 3.53°/100m and 1.33°/100m. On the relatively simpler KITTI dataset, which presents greater elevation variation and challenges the 3-DoF estimation and ground plane assumption, our method also performs competitively. Fig.~\ref{fig:ATE} shows the ATE metrics under SE(3) and Sim(3) alignment and their composite metric, illustrating the scale drift of different methods. These findings confirm that BEV representation enhances odometry accuracy and reduces scale drift for ground vehicles, with our interpretable backend further boosting performance.}

Our main contributions are as follows:
\begin{itemize}
    \item We propose BEV-DWPVO, a novel MVO framework that uses \textcolor{black}{a unified, metric-scaled} BEV representation to reduce scale drift and simplify pose estimation to 3-DoF for ground vehicles, requiring only pose supervision.
    \item Our pipeline extracts and matches keypoints on BEV feature maps, using a differentiable weighted Procrustes solver for pose estimation, outperforming other BEV-based methods in interpretability and scale consistency.
    \item We validate the effectiveness of BEV-DWPVO on the NCLT, Oxford, \textcolor{black}{and KITTI} datasets, demonstrating superior scale consistency and pose accuracy.
\end{itemize}

\section{Related Work}

\subsection{Traditional Methods}

Traditional MVO methods are highly interpretable. LSD-SLAM \cite{engel2014lsd} uses pixel intensities for pose estimation, sliding window optimization for depth maps, and loop closure. DSO \cite{engel2017direct} improves photometric calibration and jointly optimizes pose and depth. ORB-SLAM \cite{campos2021orb} relies on ORB features for matching and pose estimation, while SVO \cite{forster2014svo} optimizes sparse features using image gradients for a balance of efficiency and accuracy. \textcolor{black}{These methods anchor the scale using initial depth estimates from the first frame. However, errors from calibration, feature mismatches, or motion blur accumulate during iterative pose updates, causing scale drift and limiting their applicability to long-distance tasks.}

\subsection{Learning-Based Methods Under Perspective View}

Learning-based methods under the perspective view generally follow two approaches: employing CNNs and MLPs for pose estimation, or using side-tasks such as depth and optical flow prediction, which can further be divided into self-supervised or auxiliary supervision methods.

Early works, such as \cite{konda2015learning} and DeepVO \cite{wang2017deepvo}, rely on feature extraction and pose regression. However, both methods face limitations in interpretability, which affects their robustness in challenging scenarios.

UnDeepVO \cite{li2018undeepvo} achieves scale-absolute odometry by using stereo depth during training. TartanVO \cite{wang2021tartanvo} incorporates optical flow supervision and a scale-invariant loss to improve generalization. Self-supervised methods, such as \cite{bian2019unsupervised} and \cite{li2020self}, rely on photometric loss and SSIM but struggle with motion blur, lighting changes, and low-texture areas.

DF-VO \cite{zhan2021df} combines depth estimation and bidirectional optical flow for 3D localization, integrating deep learning with multi-view geometry. DROID-SLAM \cite{teed2021droid} estimates motion through feature extraction and inter-frame correlation, using a differentiable Dense Bundle Adjustment (DBA) layer for joint optimization of camera poses and 3D structure, significantly improving scale consistency and robustness. DPVO \cite{teed2024deep} builds on DROID-SLAM with sparse feature tracking, further enhancing efficiency.

\textcolor{black}{These methods anchor the scale using neural network parameters and depth estimation. In contrast, optical flow, which does not have a direct correspondence to the metric scale, is typically used as an auxiliary source. However, anchoring the scale directly through neural network parameters suffers from limited generalization due to the data-driven, black-box nature of the approach. Depth-based scale anchoring under a perspective view often requires additional tasks, and achieving stable performance usually necessitates extra supervision.}

\subsection{Learning-Based Methods Under BEV Representation}

Recently, some odometry methods employing BEV representation have emerged, but they primarily rely on segmentation results for pose estimation rather than directly performing the odometry task.

BEV-SLAM \cite{ross2022bev} generates BEV semantic segmentation maps using CNNs and aligns them with existing map features for pose estimation. BEV-Locator \cite{zhang2022bev} employs a BEV encoder to convert perspective images and matches semantic map elements using a cross-modal Transformer. OCC-VO \cite{li2023occ} transforms multi-camera images into 3D semantic occupancy clouds and registers them with a global map. However, these methods rely heavily on semantic segmentation, which incurs high annotation costs and limits end-to-end optimization with pose supervision, making their performance dependent on segmentation accuracy.

\textcolor{black}{Unlike traditional methods prone to scale drift and learning-based approaches requiring extra supervision and complex designs, our method fully exploits the BEV representation by directly anchoring scale to its unified, metric-scaled nature. By simplifying 6-DoF pose estimation to the primary 3-DoF motions of ground vehicles, our method enables the network to learn scale information end-to-end using only pose supervision, achieving robust scale consistency without relying on any side-tasks.}

\section{Method}

\begin{figure}[t]
\centering
\includegraphics[width=0.48\textwidth]{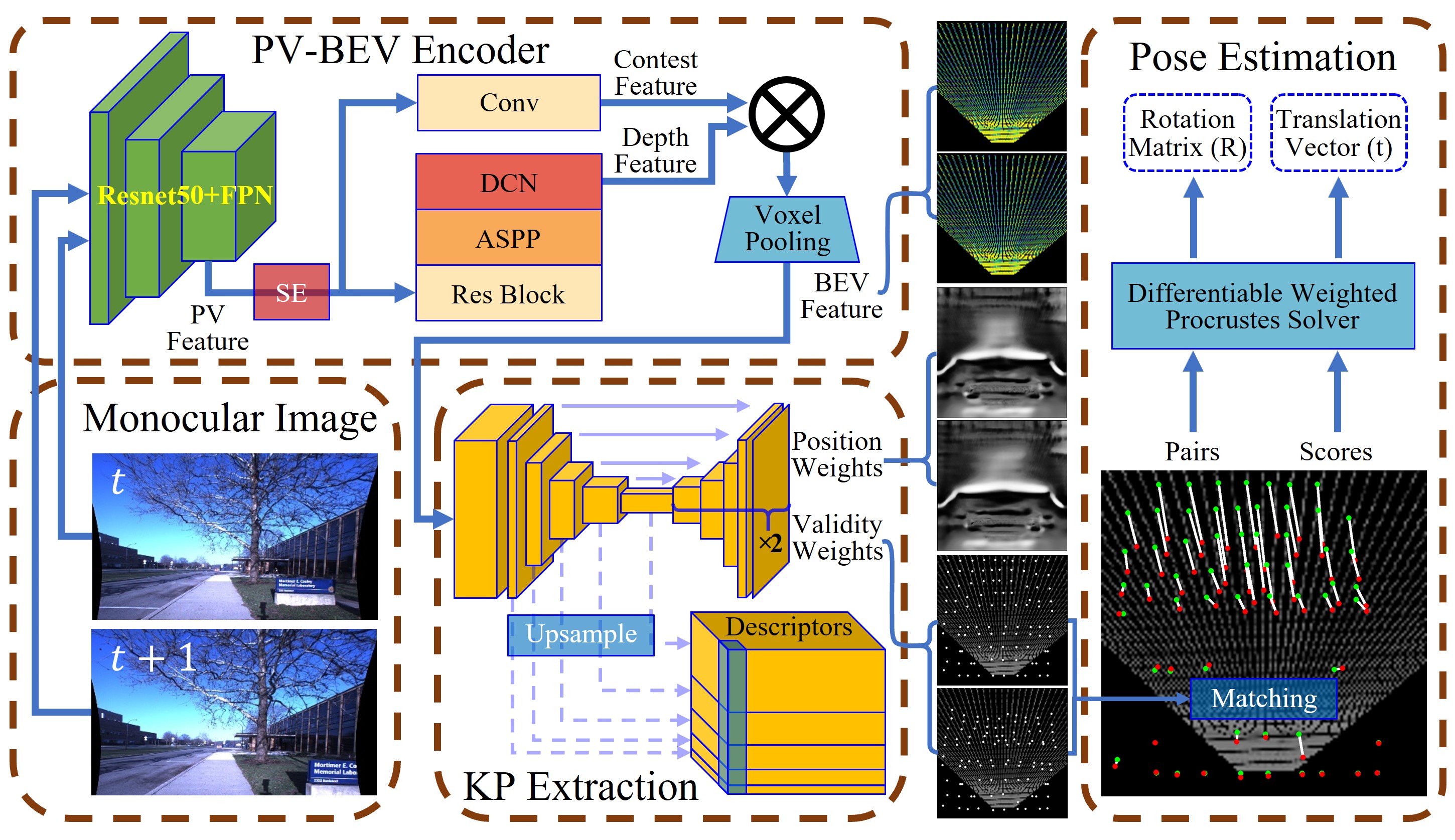}
\caption{Overview of the proposed framework.}
\label{fig:methods}
\vspace{-0.5cm}
\end{figure}

\textcolor{black}{In ground vehicle applications, the local ground plane assumption generally holds because vehicle movement mainly involves translation along the x and y axes and rotation around the z-axis. This assumption allows us to simplify pose estimation from 6-DoF to 3-DoF, enabling the method to focus on the main motion components while reducing computational complexity and minimizing noise from unnecessary movements.}

Building on this advantage, we propose BEV-DWPVO, a monocular visual odometry method that utilizes BEV-based keypoint extraction and matching, followed by an interpretable solver for relative pose estimation. The system is trained exclusively with pose supervision.

In the following sections, we detail each component of the system and explain the implementation steps. The system framework is shown in Fig.~\ref{fig:methods}.

\subsection{Visual PV-BEV Encoder}

We choose the LSS architecture \cite{philion2020lift} to extract 3D spatial features from perspective-view images and map them to BEV. LSS predicts depth distributions for scale-consistent BEV projections and incorporates camera intrinsics and extrinsics, enhancing interpretability and supporting multi-camera setups when extrinsics are precise.

We employ a ResNet-50 backbone \cite{he2016deep} with a Feature Pyramid Network (FPN) \cite{lin2017feature} to extract multi-scale feature maps \( \mathbf{F}_{\text{IF}} \) from input monocular images, with dimensions \textcolor{black}{\( C_{\text{IF}} \times H_{\text{PV}} \times W_{\text{PV}} \).}

Two networks further refine these features to predict depth and image maps, outputting \textcolor{black}{\( \mathbf{F}_{\text{PV}_{C}} \in \mathbb{R}^{C_{\text{PV}} \times H_{\text{PV}} \times W_{\text{PV}}} \) and \( \mathbf{F}_{\text{PV}_{D}} \in \mathbb{R}^{D_{\text{PV}} \times H_{\text{PV}} \times W_{\text{PV}}} \),} where depth is represented by a \textcolor{black}{\( D_{\text{PV}} \)-dimensional} distribution vector. \( \mathbf{F}_{\text{PV}_{C}} \) and \( \mathbf{F}_{\text{PV}_{D}} \) are expanded to \textcolor{black}{\( C_{\text{PV}} \times 1 \times H_{\text{PV}} \times W_{\text{PV}} \) and \( 1 \times D_{\text{PV}} \times H_{\text{PV}} \times W_{\text{PV}} \), referred to as \( \mathbf{F}'_{\text{PV}_{C}} \) and \( \mathbf{F}'_{\text{PV}_{D}} \), respectively.} Element-wise multiplication across channel and depth dimensions yields:
\textcolor{black}{\begin{equation}
\mathbf{F}_{\text{PV}_{\text{multi}}} = \mathbf{F}'_{\text{PV}_{C}} \odot \mathbf{F}'_{\text{PV}_{D}},
\end{equation}
}where \textcolor{black}{\( \mathbf{F}_{\text{PV}_{\text{multi}}} \in \mathbb{R}^{C_{\text{PV}} \times D_{\text{PV}} \times H_{\text{PV}} \times W_{\text{PV}}} \) combines features across channels and depth for each pixel.}

Finally, we project the 4D feature map onto a 2D BEV plane by compressing the height dimension. Features at different depths are mapped to 3D space as voxels, transformed to the vehicle coordinate system using camera intrinsics and extrinsics, and accumulated onto the 2D BEV plane. Voxel pooling aggregates these features, forming the final BEV feature map \textcolor{black}{\( \mathbf{F}_{\text{BEV}} \in \mathbb{R}^{C_{\text{BEV}} \times H \times W} \),} where \( C_{\text{BEV}} \) is the number of BEV feature channels\textcolor{black}{, and is equal to \( C_{\text{PV}} \)}.

\subsection{BEV Keypoint Extraction}

We employ a UNet \cite{ronneberger2015u} style encoder-decoder network to predict keypoint position weights \( \mathbf{W}_{\text{pos}} \), keypoint validity weights \( \mathbf{W}_{\text{valid}} \), and keypoint descriptors \( \mathbf{D}_{\text{key}} \) at the same resolution as the BEV feature map.

The encoder consists of four downsampling stages, progressively transforming the BEV feature map from \textcolor{black}{\( C_{\text{BEV}} \times H \times W \)} to \( 128 \times \frac{H}{16} \times \frac{W}{16} \).

In the decoding phase, both the \( \mathbf{W}_{\text{pos}} \) and \( \mathbf{W}_{\text{valid}} \) branches use four upsampling stages. These branches upsample and concatenate feature maps from \( 128 \times \frac{H}{16} \times \frac{W}{16} \) to \( 8 \times H \times W \), producing final weights with dimensions \( 1 \times H \times W \). Feature maps from each downsampling stage are also interpolated back to the original size and concatenated to form the keypoint descriptor map \( \mathbf{D}_{\text{key}} \) with dimensions \( 248 \times H \times W \), providing multi-scale descriptions for keypoints.

\subsection{Keypoint Matching and Weight Calculation}

After obtaining the keypoint position weights \( \mathbf{W}_{\text{pos}} \), keypoint validity weights \( \mathbf{W}_{\text{valid}} \), and keypoint descriptors \( \mathbf{D}_{\text{key}} \), \textcolor{black}{we perform keypoint extraction, matching, and weight calculation between two BEV feature maps. To enhance localization robustness and reduce pose estimation errors, the first BEV feature map is divided into \( h \times w \) non-overlapping blocks. This division ensures independent feature consideration across regions, preventing densely populated areas from dominating. A candidate keypoint's position \( \mathbf{P}_{\text{candidate}}^{(\text{block}_n)} \) is computed for each block from the first BEV feature map as follows:}
\textcolor{black}{
\begin{equation}
\label{eq:P1}
\mathbf{P}_{\text{candidate}}^{(\text{block}_n)} = \sum_{i \in \text{block}_n} \frac{\exp(\mathbf{W}_{\text{pos}}^{(i)})}{\sum_{j \in \text{block}_n} \exp(\mathbf{W}_{\text{pos}}^{(j)})} \cdot \mathbf{P}_{\text{orig}}^{(i)}.
\end{equation}
}

\textcolor{black}{Here, \( \mathbf{P}_{\text{orig}}^{(i)} \) represents the original coordinate of the \( i \)-th point in the block. The softmax weights from \( \mathbf{W}_{\text{pos}}^{(i)} \) provide a weighted sum of \( \mathbf{P}_{\text{orig}}^{(i)} \), ensuring differentiable keypoint selection. Since candidate keypoint positions may not align with integer pixel locations, bilinear interpolation on the keypoint descriptor map and validity weight map is used to obtain descriptors and weights for the candidate keypoints.}

\textcolor{black}{Next, keypoints from the two BEV feature maps are matched. Keypoint descriptors from the first map are normalized, and tensor multiplication with dense descriptors from the second map yields a similarity matrix with dimensions \( B \times hw \times HW \), where \( B \) is the batch size, \( hw \) is the number of blocks in the first map, and \( HW \) is the number of elements in the dense descriptors of the second map.}

\textcolor{black}{To leverage BEV’s scale properties and typical inter-frame motion, we apply a mask to zero out similarity values beyond a specified range around each keypoint. This mask range, defined as half the resolution of the BEV feature map, accounts for the maximum expected movement of corresponding keypoints between frames. This design choice ensures robustness in real-world applications without relying on dataset-specific parameter tuning.}

The similarity scores are scaled employing a temperature-weighted softmax \cite{hinton2015distilling} function:
\textcolor{black}{
\begin{equation}
\mathbf{M}_{p_1 p_2} = \frac{\exp\left(\frac{\mathbf{D}_{\text{key1}}^{p_1} \cdot \mathbf{D}_{\text{key2}}^{p_2}}{\tau}\right)}{\sum_{k \in \text{dense2}} \exp\left(\frac{\mathbf{D}_{\text{key1}}^{p_1} \cdot \mathbf{D}_{\text{key2}}^{k}}{\tau}\right)}.
\end{equation}
}

\textcolor{black}{Here, \( \mathbf{M}_{p_1 p_2} \) represents the softmax-normalized similarity score between position \( p_1 \) on the first map and \( p_2 \) on the second map. Each keypoint descriptor \( \mathbf{D}_{\text{key}} \) (length 248, as noted in the \textit{BEV Keypoint Extraction} subsection) is compared through the dot product of descriptors at these positions. The temperature parameter \( \tau \), set to 0.01 in our experiments, adjusts the sharpness of the softmax output.}

\textcolor{black}{The matching coordinates of the keypoints from the first map in the second map can be computed as follows:
\textcolor{black}{
\begin{equation}
\label{eq:P2}
\mathbf{P}_{\text{match}}^{(\text{block}_n)} = \sum_{k \in \text{dense2}} \mathbf{M}_{\mathbf{P}_{\text{candidate}}^{(\text{block}_n)}, k} \cdot \mathbf{P}_{\text{orig}}^{(k)},
\end{equation}
}
where \( \mathbf{P}_{\text{match}}^{(\text{block}_n)} \) represents the position in the second map corresponding to the keypoint in block \( n \) from the first map. \textcolor{black}{The similarity matrix \( \mathbf{M}_{\smash{\mathbf{P}_{\text{candidate}}^{(\text{block}_n)}}, k} \)} measures similarity between this keypoint and all possible positions \( k \) in the second map, with \( \mathbf{P}_{\text{orig}}^{(k)} \) as the original coordinates at \( k \).}

The final keypoint pair score considers both the similarity of the keypoint descriptors and the keypoint validity weights:
\textcolor{black}{\begin{equation}
\label{eq:S_final}
\mathbf{S}_{\text{final}}^{(\mathbf{P}_c, \mathbf{P}_m)} = \mathbf{M}_{\mathbf{P}_c \mathbf{P}_m} \cdot \mathbf{W}_{\text{valid}}^{(\mathbf{P}_c)} \cdot \mathbf{W}_{\text{valid}}^{(\mathbf{P}_m)},
\end{equation}
where \( \mathbf{S}_{\text{final}}^{(\mathbf{P}_c, \mathbf{P}_m)} \) is the final score for the keypoint pair. \( \mathbf{P}_c \) is the shorthand for \( \mathbf{P}_{\text{candidate}}^{(\text{block}_n)} \), and \( \mathbf{P}_m \) is the shorthand for \( \mathbf{P}_{\text{match}}^{(\text{block}_n)} \). \( \mathbf{M}_{\mathbf{P}_c \mathbf{P}_m} \) is the similarity score between the two keypoints, while \( \mathbf{W}_{\text{valid}}^{(\mathbf{P}_c)} \) and \( \mathbf{W}_{\text{valid}}^{(\mathbf{P}_m)} \) are the validity weights for the keypoints in the first and second maps, respectively. This equation ensures that the final matching score reflects both the descriptor similarity and the confidence in the keypoint positions.}

\subsection{Differentiable Weighted Procrustes Solver}

The differentiable weighted Procrustes solver \cite{gower1975generalized}, based on the Weighted Singular Value Decomposition (SVD) algorithm \cite{chen2017weighted}, calculates the optimal rigid transformation (rotation and translation) between two sets of keypoints. This solver supports gradient backpropagation, essential for deep learning, and maintains interpretability. Leveraging BEV's uniform scale properties, where each pixel in the top-down view corresponds to a fixed real-world distance, it effectively handles pose estimation in BEV space.

\textcolor{black}{The solver takes as input the matched keypoint pairs \( \mathbf{P}_{\text{candidate}}^{(\text{block}_n)} \) from the first BEV feature map and corresponding points \( \mathbf{P}_{\text{match}}^{(\text{block}_n)} \) from the second map, along with their final matching scores \( \mathbf{S}_{\text{final}}^{(\mathbf{P}_c, \mathbf{P}_m)} \), as detailed in Equations~\eqref{eq:P1}, \eqref{eq:P2}, and \eqref{eq:S_final}. To prepare for SVD, we convert pixel coordinates to real-world scale, obtaining \( \mathbf{P}_{\text{candidate}}^{(\text{block}_n)'} \) and \( \mathbf{P}_{\text{match}}^{(\text{block}_n)'} \).}

\textcolor{black}{Let \( w_n \) represent the final matching score for the keypoint pair in block \( n \), given by:
\begin{equation}
w_n = \mathbf{S}_{\text{final}}^{(\mathbf{P}_c^{(\text{block}_n)}, \mathbf{P}_m^{(\text{block}_n)})}.
\end{equation}
}

\textcolor{black}{The weighted centroids are then computed as:
\begin{align}
\overline{\mathbf{p}}_{\text{c}} &= \frac{\sum_{n=1}^{hw} w_n \mathbf{P}_{\text{candidate}}^{(\text{block}_n)'}}{\sum_{n=1}^{hw} w_n}, \\
\overline{\mathbf{p}}_{\text{m}} &= \frac{\sum_{n=1}^{hw} w_n \mathbf{P}_{\text{match}}^{(\text{block}_n)'}}{\sum_{n=1}^{hw} w_n},
\end{align}
where \( \overline{\mathbf{p}}_{\text{c}} \) and \( \overline{\mathbf{p}}_{\text{m}} \) are the weighted centroids of the keypoints in the first and second BEV feature maps, respectively.}

\textcolor{black}{We then subtract these centroids from the keypoint coordinates to obtain centered coordinates. Using these, we compute the weighted covariance matrix \( \mathbf{W}_{\text{cov}} \) as:
\begin{equation}
\mathbf{W}_{\text{cov}} = \frac{\sum_{n=1}^{hw} w_n \left( \mathbf{P}_{\text{candidate}}^{(\text{block}_n)'} - \overline{\mathbf{p}}_{\text{c}} \right) \left( \mathbf{P}_{\text{match}}^{(\text{block}_n)'} - \overline{\mathbf{p}}_{\text{m}} \right)^{\mathsf{T}}}{\sum_{n=1}^{hw} w_n}.
\end{equation}
}

We perform SVD on this covariance matrix \textcolor{black}{\( \mathbf{W}_{\text{cov}} \)} to obtain the matrices \( \mathbf{U} \), \( \mathbf{S} \), and \textcolor{black}{\( \mathbf{V}^{\mathsf{T}} \)}. Subsequently, we compute the rotation matrix \( \mathbf{R} \) by ensuring the determinant of the product \textcolor{black}{\( \mathbf{U} \mathbf{V}^{\mathsf{T}} \)} equals one, to meet the requirements of a proper rotation matrix:
\textcolor{black}{\begin{equation}
\mathbf{R} = \mathbf{U} \cdot \text{diag}(1, 1, \det(\mathbf{U} \mathbf{V}^{\mathsf{T}})) \cdot \mathbf{V}^{\mathsf{T}}.
\end{equation}
}

The translation vector \( \mathbf{t} \) is calculated by aligning the keypoints from the first frame to those in the second frame employing the rotation matrix \( \mathbf{R} \):
\textcolor{black}{\begin{equation}
\mathbf{t} = \overline{\mathbf{p}}_{\text{m}} - \mathbf{R} \cdot \overline{\mathbf{p}}_{\text{c}}.
\end{equation}
}

\subsection{Loss Function and Training Strategy}

The supervised pose loss \( L_{\text{pose}} \) is defined as the sum of the \( L_1 \) loss for both translation and rotation:
\[
L_{\text{pose}} = \left| t_{\text{pred},x} - t_{\text{gt},x} \right| + \left| t_{\text{pred},y} - t_{\text{gt},y} \right| + \alpha \cdot \left| \theta_{\text{pred}} - \theta_{\text{gt}} \right|,
\]
where \( \alpha = 10 \) is used to balance the translation and rotation losses, as rotation errors affect translation.

To enhance network convergence, we apply two strategies:

\textbf{Global Keypoint Pre-training}: For the first n epochs, all keypoint validity weights are set to 1, ensuring uniform attention across the BEV map and stabilizing optimization.

\textbf{Guided Convergence for Translation}: For the first n epochs, the ground truth rotation matrix is used to align the second frame’s centroid with the first frame’s coordinate system, reducing translation errors. Thereafter, the calculated rotation matrix is used.

\section{Experiments}

\subsection{Experimental Setup}

\subsubsection{Datasets}

\textbf{NCLT Dataset} \cite{carlevaris2016university}: This dataset features high jitter, significant lighting variations, and motion blur, making it suitable for evaluating performance in challenging visual conditions. Training sequences cover midday and afternoon under various weather conditions, while testing sequences include unseen times of day, weather, and seasonal features to assess generalization.

\textbf{Oxford Radar RobotCar Dataset} \cite{maddern20171}: This dataset consists of complex urban driving scenarios with varying numbers and sizes of dynamic background traffic participants, making it well-suited as a challenging urban testbed. Training sequences represent typical conditions, while testing sequences contain strong illumination conditions not seen during training.

\textcolor{black}{\textbf{KITTI Odometry Dataset} \cite{geiger2013vision}: This dataset is widely used for odometry comparison, with many methods providing pretrained models and published results, which enable consistent benchmarking. The training sequences and testing sequences include significant vertical changes, with sequence 09 reaching 38m and sequence 10 reaching 24m. These variations challenge our 3-DoF odometry method, which assumes a flat ground plane.}

\textcolor{black}{Across all datasets, we apply the same hyperparameters to avoid dataset-specific tuning. The BEV grid is set to $256 \times 256$ with a $0.4\,\text{m}$ resolution to accommodate NCLT’s low speed and jitter, Oxford’s urban complexity, and KITTI’s high-speed open scenes.}

\subsubsection{Baseline Setting}
\textcolor{black}{We compare our approach against three categories of baselines, each targeting specific aspects of our design:}

\textcolor{black}{\textbf{PV-based Baselines}: This category includes ORB-SLAM3 \cite{campos2021orb} as a traditional method, and DeepVO \cite{wang2017deepvo}, TartanVO \cite{wang2021tartanvo}, DF-VO \cite{zhan2021df}, DROID-SLAM \cite{teed2021droid}, and DPVO \cite{teed2024deep} as learning-based methods under the perspective view.}

\textcolor{black}{\textbf{BEV-based Baselines}: To investigate the impact of interpretable components such as the keypoint matching module, weight calculation module and the weighted Procrustes solver, we design three BEV-based methods:
\begin{itemize}
    \item \textbf{BEV + CNNs + MLPs}: A simple approach using CNNs and MLPs for direct feature extraction and pose regression, without keypoint-based matching.
    \item \textbf{BEV + Global Correlation + MLPs}: Computes global correlation across BEV feature maps for motion estimation but lacks spatial constraints.
    \item \textbf{BEV + Local Correlation + MLPs}: Adds spatial constraints to global correlation by focusing on local regions, improving inter-frame estimation precision.
\end{itemize}
}

\textcolor{black}{The aim of these comparisons is twofold. 
First, to validate the benefits of using a unified, metric-scaled BEV representation to simplify pose estimation from 6-DoF to 3-DoF. This design uses pose supervision to train depth-aware feature distributions, effectively anchoring scale without relying on additional tasks such as depth regression or segmentation.
Second, to assess whether interpretable designs can guide intermediate layers toward better scale anchoring and key feature capture, thereby further improving overall performance and reducing scale drift.}

\begin{figure}[t]
\centering
\includegraphics[width=0.48\textwidth]{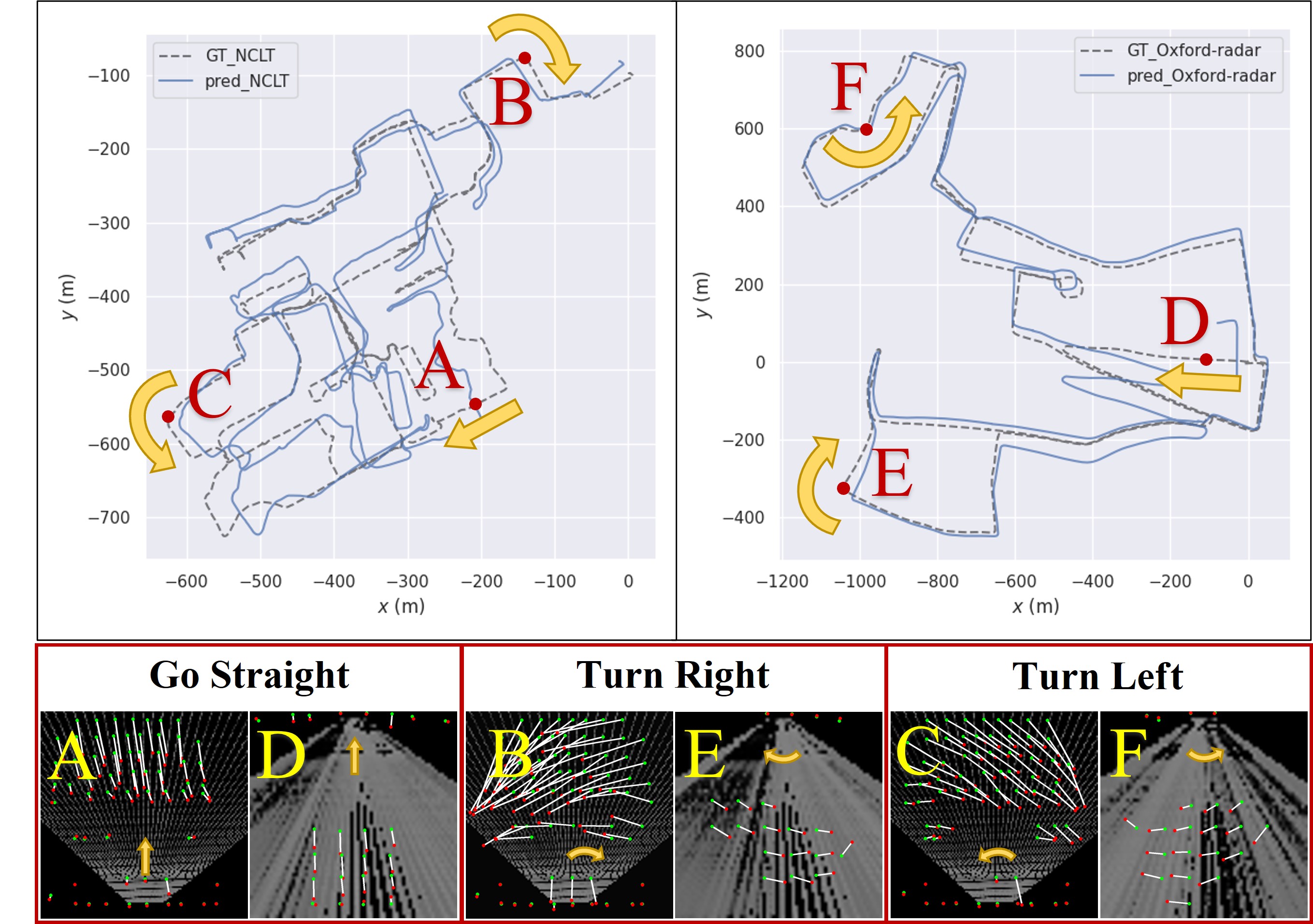}
\caption{Trajectory and keypoint matching visualization. The NCLT dataset uses a forward monocular camera, while the Oxford dataset uses a rear monocular camera. Keypoint-matching images are cropped to show relevant areas within the field of view.}
\label{fig:case_study_fig}
\vspace{-0.6cm}
\end{figure}

\renewcommand{\arraystretch}{1.2} 

\begin{table*}[t]
    \centering
    \caption{PERFORMANCE COMPARISON OF DIFFERENT METHODS ON TWO DATASETS}
    \begin{adjustbox}{max width=\textwidth}
    \begin{tabular}{
            >{\centering\arraybackslash}p{2.2cm}
            >{\centering\arraybackslash}p{2.0cm}
            |>{\centering\arraybackslash}p{1.05cm}
            >{\centering\arraybackslash}p{1.05cm}
            >{\centering\arraybackslash}p{1.05cm}
            >{\centering\arraybackslash}p{1.05cm}
            >{\centering\arraybackslash}p{1.05cm}
            |>{\centering\arraybackslash}p{1.05cm}
            >{\centering\arraybackslash}p{1.05cm}
            >{\centering\arraybackslash}p{1.05cm}
            >{\centering\arraybackslash}p{1.05cm}
            >{\centering\arraybackslash}p{1.05cm}
        }
        \toprule
        \multirow{2.5}{*}{\textbf{Methods}} & \multirow{2.5}{*}{\textbf{\diagbox[width=6em]{Metric}{Seq}}} & \multicolumn{5}{c}{\textbf{NCLT}} & \multicolumn{5}{c}{\textbf{Oxford}} \\
        \cmidrule(lr){3-7} \cmidrule(lr){8-12}
        &  & \textbf{12-02-02} & \textbf{12-02-19} & \textbf{12-03-17} & \textbf{12-08-20} & \textbf{\textit{Average}} & \textbf{01-11-12} & \textbf{01-15-13} & \textbf{01-16-14} & \textbf{01-17-12} & \textbf{\textit{Average}} \\
        \midrule
        \multirow{2}{*}{\makecell[c]{ORB-SLAM3 \cite{campos2021orb}}} & RTE** (\%) & / & / & / & / & / & 65.22 & 952.41 & 191.75 & 77.44 & \textit{321.70} \\
                                              & RRE** (°/100m) & / & / & / & / & / & 19.83 & 10.81 & 16.17 & 19.31 & \textit{16.53} \\
        \hline
        \multirow{2}{*}{\makecell[c]{DeepVO \cite{wang2017deepvo}}} & RTE** (\%) & 31.67 & 18.55 & 21.23 & 25.07 & \textit{24.13} & 25.14 & 27.45 & 31.75 & 37.76 & \textit{30.52} \\
                                 & RRE** (°/100m) & 14.54 & 8.11 & 9.66 & 13.02 & \textit{11.33} & 7.87 & 8.14 & 11.71 & 12.37 & \textit{10.02} \\
        \hline
        \multirow{2}{*}{\makecell[c]{TartanVO \cite{wang2021tartanvo} \\ (Pretrained Model)}} & RTE** (\%) & 47.48 & 50.97 & 48.10 & 52.23 & \textit{49.69} & 74.43 & 83.35 & 75.51 & 72.53 & \textit{76.45} \\
                                   & RRE** (°/100m) & 39.51 & 38.48 & 39.21 & 37.68 & \textit{38.72} & 32.56 & 25.68 & 31.63 & 33.23 & \textit{30.77} \\
        \hline
        \multirow{2}{*}{\makecell[c]{DF-VO \cite{zhan2021df} \\ (F. Model)}} & RTE** (\%) & / & / & 41.03 & 89.44 & \textit{65.23} & 26.24 & 28.26 & 35.87 & 42.48 & \textit{33.21} \\
                                                      & RRE** (°/100m) & / & / & 25.52 & 27.81 & \textit{26.66} & 2.03 & 2.34 & 1.85 & 2.81 & \textit{2.25} \\
        \hline
        \multirow{2}{*}{\makecell[c]{DROID-SLAM \cite{teed2021droid} \\ (Pretrained Model)}} & RTE** (\%) & 34.07 & 110.18 & 44.17 & 41.80 & \textit{57.55} & 35.18 & 136.58 & 133.31 & 25.82 & \textit{82.72} \\
                                    & RRE** (°/100m) & 14.23 & 10.50 & 10.67 & 12.04 & \textit{11.86} & 1.89 & 1.43 & 1.99 & 3.47 & \textit{2.19} \\
        \hline
        \multirow{2}{*}{\textcolor{black}{\makecell[c]{DPVO \cite{teed2024deep} \\ (Pretrained Model)}}}  
        & \textcolor{black}{RTE** (\%)} & \textcolor{black}{/} & \textcolor{black}{41.45} & \textcolor{black}{37.72} & \textcolor{black}{78.87} & \textcolor{black}{\textit{52.68}} & \textcolor{black}{96.53} & \textcolor{black}{98.41} & \textcolor{black}{102.10} & \textcolor{black}{155.51} & \textcolor{black}{\textit{113.14}} \\
        & \textcolor{black}{RRE** (°/100m)} & \textcolor{black}{/} & \textcolor{black}{9.74} & \textcolor{black}{14.38} & \textcolor{black}{16.70} & \textcolor{black}{\textit{13.61}} & \textcolor{black}{29.26} & \textcolor{black}{28.00} & \textcolor{black}{28.37} & \textcolor{black}{29.14} & \textcolor{black}{\textit{28.69}} \\
        \hline
        \multirow{2}{*}{\makecell[c]{BEV \\ (CNNs+MLPs)}} & RTE* (\%) & 21.36 & 18.77 & 13.95 & 16.14 & \textit{17.55} & 8.68 & 10.43 & \underline{10.99} & 16.97 & \textit{11.76} \\
                                      & RRE* (°/100m) & 10.54 & 7.72 & 6.61 & 7.50 & \textit{8.09} & 2.24 & 2.72 & 2.03 & 5.66 & \textit{3.16} \\
        \hline
        \multirow{2}{*}{\makecell[c]{BEV \\ (Global-Corr)}} & RTE* (\%) & 13.54 & 10.94 & \textbf{7.84} & 16.38 & \textit{12.17} & 11.18 & 9.34 & 15.26 & 11.07 & \textit{11.71} \\
                                              & RRE* (°/100m) & 6.44 & 5.36 & \underline{3.60} & 7.89 & \textit{5.82} & 2.41 & 2.24 & 3.15 & 3.06 & \textit{2.71} \\
        \hline
        \multirow{2}{*}{\makecell[c]{BEV \\ (Local-Corr)}} & RTE* (\%) & \underline{10.89} & \textbf{6.27} & 8.77 & \underline{13.89} & \underline{\textit{9.95}} & \underline{7.21} & \underline{7.90} & 11.55 & \textbf{8.02} & \underline{\textit{8.67}} \\
                                             & RRE* (°/100m) & \underline{4.74} & \textbf{2.47} & 3.73 & \underline{6.31} & \underline{\textit{4.31}} & \underline{1.38} & \underline{1.89} & \underline{1.14} & \underline{2.55} & \underline{\textit{1.74}} \\
        \hline
        \multirow{2}{*}{\textcolor{black}{\makecell[c]{BEV-DWPVO \\ (Ours)}}} 
        & \textcolor{black}{RTE* (\%)} & \textcolor{black}{\textbf{8.32}} & \textcolor{black}{\underline{7.52}} & \textcolor{black}{\underline{8.20}} & \textcolor{black}{\textbf{10.63}} & \textcolor{black}{\textbf{\textit{8.67}}} & \textcolor{black}{\textbf{4.87}} & \textcolor{black}{\textbf{5.26}} & \textcolor{black}{\textbf{7.89}} & \textcolor{black}{\underline{9.67}} & \textcolor{black}{\textbf{\textit{6.92}}} \\
        & \textcolor{black}{RRE* (°/100m)} & \textcolor{black}{\textbf{3.14}} & \textcolor{black}{\underline{3.08}} & \textcolor{black}{\textbf{3.57}} & \textcolor{black}{\textbf{4.35}} & \textcolor{black}{\textbf{\textit{3.53}}} & \textcolor{black}{\textbf{1.02}} & \textcolor{black}{\textbf{1.08}} & \textcolor{black}{\textbf{1.07}} & \textcolor{black}{\textbf{2.15}} & \textcolor{black}{\textbf{\textit{1.33}}} \\
        \hline
        \bottomrule
    \end{tabular}
    
    \end{adjustbox}
    \label{table:main_table}
    \captionsetup{justification=raggedright, singlelinecheck=false}
    \caption*{\footnotesize{\textcolor{black}{\textbf{Pretrained Model}: Results obtained using pretrained weights;}
    \textbf{F. Model}: Use foundation model for optical flow and monocular depth estimation.\\
    * Aligned using SE(3). \\
    ** Scaled by the first 10m's ground truth and aligned using SE(3).}}
    \captionsetup{justification=centering, singlelinecheck=true}
\vspace{-0.6cm}
\end{table*}

\renewcommand{\arraystretch}{1.0}

\renewcommand{\arraystretch}{1.2} 

\begin{table}[t]
    \centering
    \caption{PERFORMANCE COMPARISON ON KITTI}
    \begin{adjustbox}{max width=\columnwidth}
    \begin{tabular}{ 
            >{\centering\arraybackslash}c
            >{\centering\arraybackslash}c
            |>{\centering\arraybackslash}c
            >{\centering\arraybackslash}c
            >{\centering\arraybackslash}c
        }
        \toprule
        \multirow{2}{*}{\textbf{Methods}} & \multirow{2}{*}{\textbf{\diagbox[width=6em]{Metric}{Seq}}} & \multirow{2}{*}{\textbf{KITTI-09}} & \multirow{2}{*}{\textbf{KITTI-10}} & \multirow{2}{*}{\textbf{Average}} \\
        & & & & \\
        \midrule
        \multirow{2}{*}{\textcolor{black}{\makecell[c]{ORB-SLAM3 \cite{campos2021orb}}}} 
            & \textcolor{black}{RTE** (\%)} & \textcolor{black}{14.89} & \textcolor{black}{8.47} & \textcolor{black}{11.68} \\
            & \textcolor{black}{RRE** (°/100m)} & \textcolor{black}{1.38} & \textcolor{black}{0.93} & \textcolor{black}{1.15} \\
        \hline
        \multirow{2}{*}{\textcolor{black}{\makecell[c]{DeepVO \cite{wang2017deepvo}}}} 
            & \textcolor{black}{RTE** (\%)} & \textcolor{black}{33.55} & \textcolor{black}{30.46} & \textcolor{black}{32.01} \\
            & \textcolor{black}{RRE** (°/100m)} & \textcolor{black}{14.31} & \textcolor{black}{14.42} & \textcolor{black}{14.37} \\
        \hline
        \multirow{2}{*}{\textcolor{black}{\makecell[c]{TartanVO \cite{wang2021tartanvo} \\ (Original Paper)}}} 
            & \textcolor{black}{RTE (\%)} & \textcolor{black}{6.00} & \textcolor{black}{6.89} & \textcolor{black}{6.45} \\
            & \textcolor{black}{RRE (°/100m)} & \textcolor{black}{3.11} & \textcolor{black}{2.73} & \textcolor{black}{2.92} \\
        \hline
        \multirow{2}{*}{\textcolor{black}{\makecell[c]{DF-VO \cite{zhan2021df} \\ (Original Paper)}}} 
            & \textcolor{black}{RTE (\%)} & \textcolor{black}{\underline{2.40}} & \textcolor{black}{\textbf{1.82}} & \textcolor{black}{\textbf{2.11}} \\
            & \textcolor{black}{RRE (°/100m)} & \textcolor{black}{\textbf{0.24}} & \textcolor{black}{\underline{0.38}} & \textcolor{black}{0.31} \\
        \hline
        \multirow{2}{*}{\textcolor{black}{\makecell[c]{DROID-SLAM \cite{teed2021droid} \\ (Pretrained Model)}}} 
            & \textcolor{black}{RTE** (\%)} & \textcolor{black}{21.01} & \textcolor{black}{18.73} & \textcolor{black}{19.87} \\
            & \textcolor{black}{RRE** (°/100m)} & \textcolor{black}{\underline{0.33}} & \textcolor{black}{\textbf{0.23}} & \textcolor{black}{\underline{0.28}} \\
        \hline
        \multirow{2}{*}{\textcolor{black}{\makecell[c]{DPVO \cite{teed2024deep} \\ (Pretrained Model)}}} 
            & \textcolor{black}{RTE** (\%)} & \textcolor{black}{17.70} & \textcolor{black}{4.47} & \textcolor{black}{11.09} \\
            & \textcolor{black}{RRE** (°/100m)} & \textcolor{black}{\textbf{0.24}} & \textcolor{black}{\textbf{0.23}} & \textcolor{black}{\textbf{0.23}} \\
        \hline
        \multirow{2}{*}{\textcolor{black}{\makecell[c]{BEV-DWPVO \\ (Ours)}}} 
            & \textcolor{black}{RTE* (\%)} & \textcolor{black}{\textbf{2.13}} & \textcolor{black}{\underline{3.24}} & \textcolor{black}{\underline{2.69}} \\
            & \textcolor{black}{RRE* (°/100m)} & \textcolor{black}{0.65} & \textcolor{black}{1.15} & \textcolor{black}{0.90} \\
    \bottomrule
    \end{tabular}
    \end{adjustbox}
    \label{table:kitti_results}
    \captionsetup{justification=raggedright, singlelinecheck=false}
    \caption*{\footnotesize{\textcolor{black}{\textbf{Original Paper}: Results as reported in the original publication;}
    \textcolor{black}{\textbf{Pretrained Model}: Results obtained using pretrained weights.}\\
    * Aligned using SE(3). \\
    ** Scaled by the first 10m's ground truth and aligned using SE(3).}}
    \captionsetup{justification=centering, singlelinecheck=true}
\vspace{-0.8cm}
\end{table}

\renewcommand{\arraystretch}{1.0}

\subsubsection{Experiments Setting}
To simulate varied speeds and motions, we randomly sample frame pairs and oversample frames with significant rotations. Testing trajectories are computed by accumulating relative poses at fixed intervals without any post-processing. \textcolor{black}{Training is performed for 50 epochs on an NVIDIA RTX 4090 GPU using the Adam optimizer, starting at a learning rate of \(1 \times 10^{-4}\) and decaying by 0.95 per epoch. The entire training process takes approximately 72 hours.}

\subsubsection{Evaluation Strategy}
We evaluate odometry accuracy using RTE, RRE, and ATE. RTE is the average translational RMSE over 100 to 800 meters, RRE is the average rotational RMSE over the same range, and ATE measures the mean translation error between predicted and ground truth poses. We also design a composite metric, \( \log_2(\text{SE(3)}/\text{Sim(3)}) \), to quantify the change in error before and after scale alignment, This metric effectively characterizes the scale consistency performance of different methods when the ATE itself is relatively small.

To ensure fairness, we scale trajectories based on the ground truth over the first 10 meters for MVO methods that lack absolute scale or rely on pre-trained models, aligning with real-world MVO usage.

To test whether side-tasks methods could leverage the latest foundation models to achieve excellent performance on datasets lacking the required supervision, we replace the bidirectional optical flow prediction and depth estimation networks of DF-VO with foundation models ZoeDepth \cite{bhat2023zoedepth} and Unimatch-Flow \cite{xu2023unifying}. Additionally, we disable global bundle adjustment in DROID-SLAM and loop closure in ORB-SLAM3. \textcolor{black}{Due to open-source limitations, we train DeepVO and BEV-based methods, while TartanVO, DROID-SLAM, and DPVO use official pre-trained models.}

\subsection{Qualitative Analysis}

The upper part of Fig.~\ref{fig:case_study_fig} shows trajectory comparisons between our method and ground truth under Sim(3) alignment, with NCLT results on the left and Oxford on the right. We highlight three example locations: straight, right turn, and left turn. In the lower part, the background of each plot is the BEV feature map of the previous frame. Red points show the keypoints in the previous frame's BEV feature map, green points show the corresponding keypoints in the next frame's BEV feature map, and white lines showing their correspondences. For clarity, Non-Maximum Suppression (NMS) is set to 0.1, displaying only matches with scores above 10\% of the highest score.

\textcolor{black}{The trajectory plots show that our method achieves high path accuracy in long-distance and complex turning scenarios on both the NCLT and Oxford datasets without any post-processing. The lower part further demonstrates our system's capability to recognize and leverage geometric patterns in the BEV feature map. By directly solving transformations from weighted matched keypoints, the differentiable weighted Procrustes solver provides a structured and interpretable approach to pose estimation. This interpretability improves the network’s capability to represent keypoints in intermediate layers and allows effective optimization with only pose supervision.}

\textcolor{black}{Overall, the results validate that our feature-based BEV method achieves reliable and accurate odometry estimation, effectively handling long-distance and complex trajectories. The interpretability of the weighted feature matching and pose estimation modules further enhances the system's ability to capture structured feature relationships and achieve robust performance across diverse scenarios.}

\subsection{Quantitative Analysis}

Table~\ref{table:main_table} and Fig.~\ref{fig:ATE} present the RTE, RRE, and ATE metrics using SE(3) and Sim(3) alignment for all methods on four test sequences from the NCLT and Oxford datasets. Our method achieves the best average performance across all metrics in both datasets.

\textcolor{black}{Compared to PV-based baselines such as DeepVO, TartanVO, DF-VO, DROID-SLAM, and DPVO, all BEV-based methods achieve significantly better RTE metrics on both the NCLT and Oxford datasets, with overall advantages in RRE metrics. These experimental results indicate that our unified, metric-scaled BEV grid enables 3-DoF estimation and leverages scale-consistent representations learned solely through pose supervision, which enhances stability and accuracy across challenging environments.}

\textcolor{black}{Among BEV-based baselines, methods using global cross-correlation for motion extraction outperform those relying solely on CNNs and MLPs. Adding spatial constraints with local cross-correlation further improves precision. Our BEV-DWPVO method achieves the best results, leveraging an interpretable, feature-based framework that combines keypoint extraction, matching, and a differentiable weighted Procrustes solver.}

\renewcommand{\arraystretch}{1.1} 

\begin{table}[t]
\centering
\caption{SCALE DRIFT ON TWO DATASETS}
\setlength{\tabcolsep}{3pt}
\begin{adjustbox}{max width=0.48\textwidth}
\begin{tabular}{>{\centering\arraybackslash}p{3cm}|
>{\centering\arraybackslash}p{2.5cm}
>{\centering\arraybackslash}p{2.5cm}
}
\toprule
\textbf{Methods} & \textbf{NCLT} & \textbf{Oxford} \\
\midrule
ORB-SLAM3 \cite{campos2021orb} & 0.6275 & 2.3164 \\
DeepVO \cite{wang2017deepvo} & 0.3090 & 0.5880 \\
TartanVO \cite{wang2021tartanvo} & 0.6785 & 1.5112 \\
DF-VO \cite{zhan2021df} & 0.5886 & 1.5331 \\
DROID-SLAM \cite{teed2021droid} & 0.8816 & 1.7768 \\
DPVO \cite{teed2024deep} & 3.9275 & 0.5536 \\
BEV (CNNs+MLPs) & 0.0887 & 0.1742 \\
BEV (Global-Corr) & 0.0771 & 0.1443 \\
BEV (Local-Corr) & \underline{0.0662} & \underline{0.1199} \\
Ours & \textbf{0.0512} & \textbf{0.0573} \\
\bottomrule
\end{tabular}
\end{adjustbox}
\label{table:mean_deviation_performance}
\vspace{-0.2cm}
\end{table}

\renewcommand{\arraystretch}{1.0} 

\begin{figure}[t]
\centering
\includegraphics[width=0.48\textwidth]{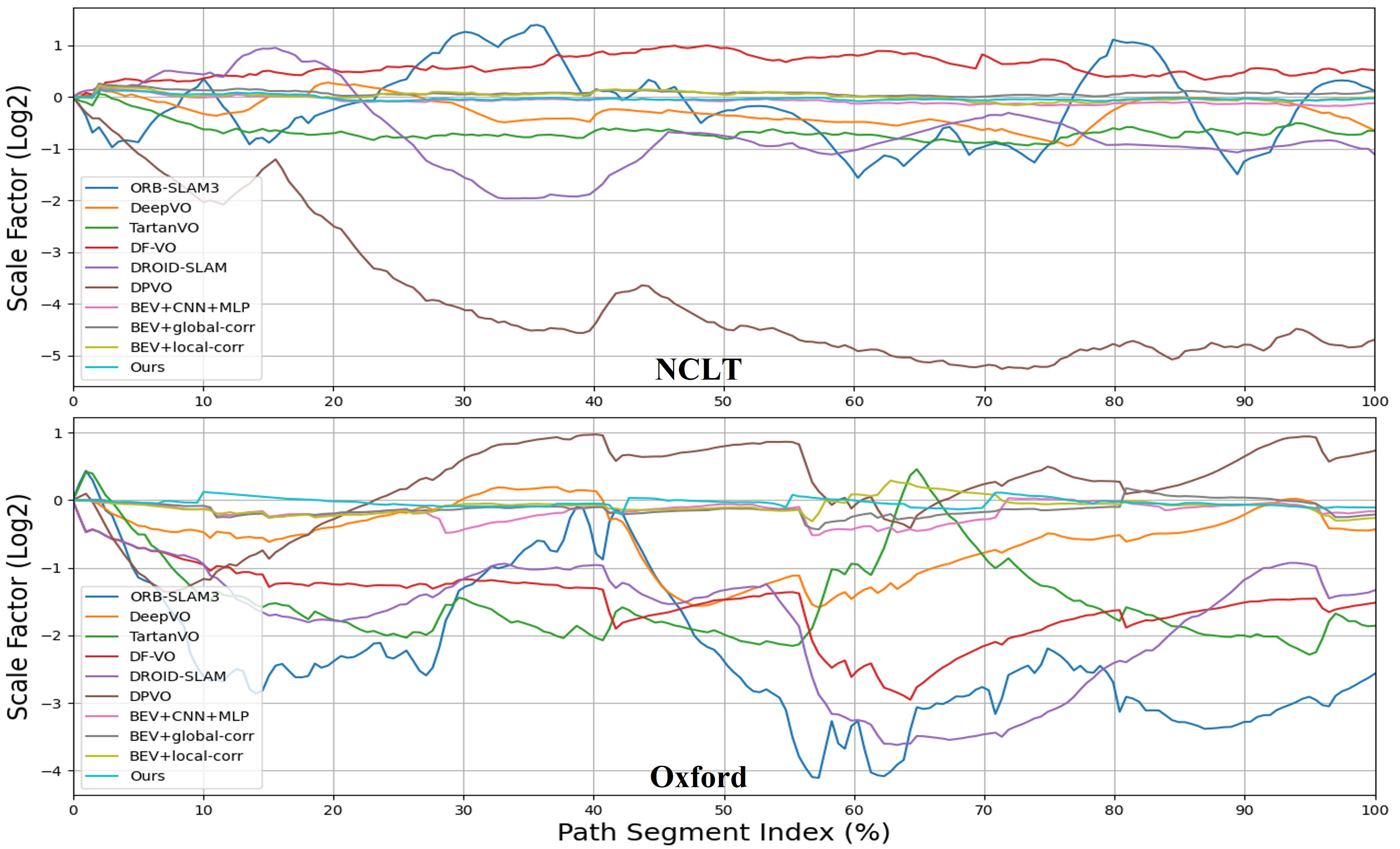}
\caption{Logarithmic scale factor variation along the path.}
\label{fig:scale_drift_fig}
\vspace{-0.6cm}
\end{figure}

\textcolor{black}{The \( \log_2(\text{SE(3)}/\text{Sim(3)}) \) boxplot in Fig.~\ref{fig:ATE} demonstrates that BEV-based methods offer superior scale consistency, with our BEV-DWPVO achieving the best performance and the lowest ATE metrics under both alignment rules. This highlights the effectiveness of using a more interpretable pipeline with BEV representations. By explicitly solving relative poses through keypoint matching and weighted optimization, our approach addresses common issues like underfitting and feature ambiguity in direct regression methods. Furthermore, the integration of geometric constraints with learned features provides structured feedback during training, improving intermediate representations and aligning them with the geometric requirements of the task.}

\textcolor{black}{Table~\ref{table:kitti_results} shows that while our method achieves competitive performance on the KITTI dataset, it also reveals a trade-off between strict geometric constraints and motion representation flexibility. By assuming a 3-DoF pose estimation under the ground plane assumption, our approach sacrifices vertical motion representation but improves scale consistency and accuracy in primary motion estimation. This trade-off is more evident in sequences with significant elevation changes. However, trained on sequences 00-08 and evaluated on sequences 09-10, our method maintains reasonable accuracy, demonstrating effective generalization within the simplified motion model. These results validate our design choice to reduce degrees of freedom for ground vehicle odometry, while acknowledging its limitations.}

\textcolor{black}{We also evaluate runtime on the KITTI dataset using an NVIDIA RTX 4090 GPU. With a batch size of 1 and input size (3, 1216, 384), our method achieves 16 FPS, 45\% GPU utilization, and 7.5GB memory usage, including data transmission delays. This demonstrates its suitability for real-time visual odometry.}

\renewcommand{\arraystretch}{1.1} 

\begin{table}[t]
\centering
\setlength{\tabcolsep}{4pt}
\caption{ABLATION STUDY ON TWO DATASETS}
\begin{adjustbox}{max width=0.48\textwidth}
\begin{tabular}{cccc|cc|cc}
\toprule
\multirow{3}{*}{\parbox{0.8cm}{\centering \textbf{UKVW}}} & \multirow{3}{*}{\parbox{0.6cm}{\centering \textbf{GKP}}} & \multirow{3}{*}{\parbox{0.6cm}{\centering \textbf{GCT}}} & \multirow{3}{*}{\parbox{1.75cm}{\centering \textbf{BEV Grid \\ Resolution \\ \& Size}}} & \multicolumn{2}{c}{\textbf{NCLT}} & \multicolumn{2}{c}{\textbf{Oxford}} \\
\cline{5-6} \cline{7-8}
& & & & \makecell{\rule{0pt}{3ex} \textbf{RTE*}\\[-2pt](\%)} & \makecell{\rule{0pt}{3ex} \textbf{RRE*}\\[-2pt](°/100m)} & \makecell{\rule{0pt}{3ex} \textbf{RTE*}\\[-2pt](\%)} & \makecell{\rule{0pt}{3ex} \textbf{RRE*}\\[-2pt](°/100m)} \\
\midrule
\textcolor{black}{$\checkmark$} & \textcolor{black}{$\checkmark$} & \textcolor{black}{$\checkmark$} & \textcolor{black}{256\&0.4} & \textcolor{black}{8.67} & \textcolor{black}{3.53} & \textcolor{black}{\underline{6.92}} & \textcolor{black}{\textbf{1.33}} \\
$\checkmark$ & $\checkmark$ & $\checkmark$ & 256\&0.2 / 128\&0.8 & \textbf{7.95} & \textbf{3.32} & \textbf{6.34} & \underline{1.50} \\
 &  & $\checkmark$ & 256\&0.2 / 128\&0.8 & 10.66 & 4.78 & 8.62 & 1.99 \\
$\checkmark$ &  & $\checkmark$ & 256\&0.2 / 128\&0.8 & 9.15 & 3.86 & \underline{6.92} & 1.71 \\
$\checkmark$ & $\checkmark$ &  & 256\&0.2 / 128\&0.8 & \underline{8.29} & \underline{3.46} & 7.02 & 1.64 \\
\bottomrule
\end{tabular}
\end{adjustbox}
\label{table:Ablation_study}
\captionsetup{justification=raggedright, singlelinecheck=false}
\caption*{\footnotesize{\textbf{UKVW}: Use Keypoint Validity Weights;
\textbf{GKP}: Global Keypoint Pre-training;
\textbf{GCT}: Guided Convergence for Translation. \\
* Aligned using SE(3).}}
\captionsetup{justification=centering, singlelinecheck=true}
\vspace{-0.8cm}
\end{table}

\renewcommand{\arraystretch}{1.0}

\subsection{Scale Consistency Analysis}

We align the scale of all PV-based methods using the ground truth over the first 10 meters, then evaluate the scale drift on one trajectory from each of the NCLT and Oxford datasets using the following formula:
\begin{equation}
D_{\text{scale}} = \frac{1}{N} \sum_{i=1}^{N} \left| \log_2\left(\frac{d_i}{d_{i}^{\text{GT}}}\right) \right|.
\end{equation}
where \( d_i \) and \( d_{i}^{\text{GT}} \) represent the estimated and ground truth displacements for segment \( i \). 
The logarithmic function standardizes proportional errors across magnitudes, while the absolute value and averaging provide a comprehensive assessment of scale drift over the entire trajectory.

\textcolor{black}{ORB-SLAM3 anchors the scale through initial motion estimates, resulting in cumulative scale drift over long trajectories. DeepVO anchors the scale with a deep network, but its limited pose estimation accuracy leads to significant errors. TartanVO, DROID-SLAM, and DPVO anchor the scales using optical flow for auxiliary training; however, the lack of inherent scale anchoring in perspective-view optical flow reduces their effectiveness in novel environments. DF-VO anchors the scale via monocular depth estimation but struggles with generalization due to its reliance on complex designs and multiple side tasks. None of these methods can anchor the scale reliably in unseen environments using only pose ground truth. These limitations are reflected in Table~\ref{table:mean_deviation_performance} and Fig.~\ref{fig:scale_drift_fig}.}

\textcolor{black}{In contrast, BEV-based methods, utilizing the ground-plane assumption, simplify 6-DoF pose estimation to 3-DoF, enabling robust scale anchoring using only pose supervision. Our approach further enhances this by explicitly solving relative poses through keypoint matching and weighted optimization, addressing common issues like feature ambiguity in direct regression methods. As a result, our method achieves the best performance in scale drift evaluations.}

\subsection{Ablation Study}

Table~\ref{table:Ablation_study} presents our ablation study evaluating key components of BEV-DWPVO. We analyze three critical aspects:

\begin{itemize}
    \item \textbf{Keypoint Validity Weights:} We add validity weights to keypoints to capture their global importance in addition to position weights. Removing these weights significantly decreases performance, showing the benefit of combining local and global weights for accurate keypoint matching.
    
    \item \textbf{Training Strategies:} We test strategies such as Global Keypoint Pre-training and Guided Convergence for Translation. These approaches improve RTE and RRE metrics by aiding early-stage convergence.

    \item \textcolor{black}{\textbf{BEV Hyperparameters:} We examine the effect of adjusting BEV grid parameters across different datasets. Results show that a finer grid resolution improves feature capture in NCLT's low-speed scenarios with motion jitter, while a coarser grid better handles Oxford's high-speed urban driving with stable vehicle motion.}
\end{itemize}

\section{Conclusion}

In this paper, we present BEV-DWPVO, a MVO system that uses BEV representation and a differentiable weighted Procrustes solver for scale-consistent pose estimation. By simplifying the task to three motion dimensions under the ground plane assumption, our method anchors scale to the depth distribution prediction network, enabling an end-to-end differentiable pipeline trained solely on pose information without extra supervision.

Experiments on the NCLT, Oxford, and KITTI datasets show that BEV-DWPVO outperforms existing MVO methods, reducing scale drift across long sequences. It also demonstrates strong generalization across different environments without dataset-specific parameter tuning, highlighting the robustness and applicability of BEV representation for ground vehicle visual odometry.

\textcolor{black}{Future work will focus on extending our framework through multi-layer BEV representation and adaptive ground plane estimation to incorporate elevation information, enabling robust handling of non-planar terrain and expanding its applicability while maintaining the advantages of our unified, metric-scaled approach.}

\addtolength{\textheight}{-12cm}   










\small
\bibliographystyle{ieeetr}
\bibliography{root}

\end{document}